\documentclass{article}

\usepackage{ijcai23}

\usepackage{times}
\usepackage{soul}
\usepackage{url}
\usepackage[hidelinks]{hyperref}
\usepackage[utf8]{inputenc}
\usepackage[small]{caption}
\usepackage{graphicx}
\usepackage{amsmath}
\usepackage{amsthm}
\usepackage{booktabs}
\usepackage{algorithm}
\usepackage{algorithmic}
\usepackage[switch]{lineno}
 \usepackage{amsmath}
 \usepackage{txfonts}
 \usepackage{graphicx}
 \usepackage{pifont}
 \usepackage{caption}
\usepackage{graphicx}
\usepackage{lipsum}
\usepackage{float} 
\usepackage{subcaption}
\usepackage{times}
\usepackage{soul}
\usepackage{url}

\title{Bright Channel Prior Attention for Multispectral Pedestrian Detection}
\author{
        {Chenhang Cui$^{1}$\thanks{Three authors contributed equally to this work.}, Jinyu Xie$^{1*}$,  Yechenhao Yang$^{1*}$}
    \affiliations
{{$^{1}$  Yingcai Honors College, University of Electronic Science
and Technology of China, Chengdu, China}}\\
  }

\begin{document}

\maketitle

\begin{abstract}
Multispectral methods have gained considerable attention due to their promising performance across various fields. However,  most existing methods cannot effectively utilize information from two modalities while optimizing time efficiency. These methods often prioritize  accuracy or time efficiency, leaving room for improvement in their performance. To this end, we propose a  new method bright channel prior attention  for enhancing pedestrian detection in low-light conditions by integrating image enhancement and detection within a unified framework. The method uses the V-channel of the HSV image of the thermal image as an attention map to trigger the unsupervised auto-encoder for visible light images, which gradually emphasizes pedestrian features across layers.
Moreover,   we utilize  unsupervised bright channel prior algorithms  to address light compensation in low light images. The proposed method includes a self-attention enhancement module and a detection module, which work together to improve object detection. An initial illumination map is estimated using the BCP,  guiding the learning of the self-attention map from  the enhancement network to obtain more informative representation focused on pedestrians. The   extensive  experiments   show effectiveness of the proposed method is demonstrated through.
\end{abstract}


\section{Introduction}
Pedestrian detection \cite{dollar2009pedestrian,dollar2011pedestrian} is a vital aspect of autonomous driving technology. Its basic task is to detect and classify pedestrians on the roadway. As autonomous driving technology evolves, unmanned vehicles must precisely sense their surrounding, particularly in pedestrian detection to guarantee safety and efficiency. As a result, the accurate and low latency detection of pedestrians has become an essential issue in autonomous driving. Nevertheless, the traditional pedestrian detection algorithm has limited effectiveness in low-light conditions. Researchers have made significant efforts to enhance the performance of pedestrian detection algorithms in low-light conditions \cite{li2021deep,wang2022sfnet}.

 Multispectral methods have gained considerable attention due to their promising performance across various fields,  such as  image super resolution \cite{thomas2008synthesis},  remote sensing \cite{berni2009thermal} and detection \cite{osorio2020deep}.
The multispectral method is a frequently employed enhancement   
traditional pedestrian detection algorithms. Traditional pedestrian detection algorithms often utilize grayscale or RGB images for detection \cite{ouyang2013joint}. Nevertheless, these images provide inadequate color and texture information, resulting in inaccurate detection of pedestrians. The multispectral method uses data from various wavelength bands to provide more color and texture information, enhancing pedestrian detection algorithms' accuracy and robustness. In addition, scholars extensively use infrared thermal images in other modes since they provide valuable contour information. Multispectral pedestrian detection algorithms utilizing visible and infrared light images have garnered increasing research attention \cite{hwang2015multispectral,cao2021handcrafted}.

Currently, many  multimodal algorithms \cite{avramidis2022enhancing,xu2022towards} employ the parallel bi-stream mode  of visible light and infrared thermal imaging. \cite{wagner2016multispectral} utilize the complementarity of visible light and infrared thermal imaging information, fusing them at a stage in the model, and then sending them to the pedestrian detection module for processing.  \cite{pei2020fast}
propose a multispectral pedestrian detector     by combining visual-optical (VIS) and IR images, utilizing a deep convolutional neural network (DCNN) with the sum fusion strategy.  
\cite{zhang2021guided}  introduce an appoach designed with inter- and intra-modality attention modules, which enables deep learning models to dynamically weigh and merge multispectral features.
Although the proposed structures can potentially improve detection accuracy, the challenge of effectively utilizing information from two modalities while optimizing time efficiency persists. To address above issues, we propose an attention-based feature fusion algorithm to direct the integration of multimodal information and pedestrian detection algorithm based on Yolov4 model to advance performance.

The main contributions of this paper are as follows:
\begin{itemize}
	\item First, we use the V-channel (brightness) of the HSV image of the thermal image  as an attention map to trigger the unsupervised auto-encoder for visible light images. This attention map gradually emphasizes pedestrian features across layers.
	\item Second, our approach utilizes advanced networks and unsupervised bright channel prior algorithms to address light compensation in low light images.
	\item Thirdly, to improve object detection, we integrate image enhancement and detection within a unified framework, allowing  the enhancement benefits to extend to detection tasks.
\end{itemize}
\section{Related Works}

\subsection{Multispectral Pedestrian Detection}
Pedestrian detection has significant practical significance in various fields, particularly in developing advanced driver assistance systems (ADAS)\cite{geronimo2009survey}  and autonomous vehicles. The capability to detect pedestrians is critical for ensuring safety in these applications. In recent years,  multispectral pedestrian detection  due to its more robust performance  in certain special environments. Traditional methods  including   SVM\cite{garcia2011multispectral} and AdaBoost \cite{freund1997decision},   achieved the good performance in  multispectral pedestrian detection. However,  recently  deep learning based methods  further improved detection performance compared to previous methods. \cite{zheng2019gfd} introduces a novel multispectral pedestrian detection algorithm called GFD-SSD, which incorporates gated fusion technology with the bidirectional feature map transformation SSD network. \cite{zhang2019cross}  proposes a novel multispectral pedestrian detection algorithm called Cross-Modality Interactive Attention Network (CIAN), designed to exploit the correlations between RGB and thermal data for improved detection accuracy.
\subsection{Low-Light Image Enhancement}
Low-light image enhancement is a technique that enhances the quality of images taken in low-quality environments. This technique improves image clarity and brightness, making the images more visually appealing and informative. \cite{he2010single}  proposes a new algorithm for single image haze removal based on the dark channel prior principle.   \cite{guo2016lime}  uses illumination map estimation to correct images captured in low-light conditions. \cite{ren2020lr3m}  proposes a novel low-light image enhancement algorithm called LR3M, which utilizes a low-rank regularized retinex model to effectively enhance low-light images with superior robustness and high image quality.

\section{Proposed Method}
The proposed method's overall architecture is shown in Fig. . The   model extends the established framework of YOLO and consists of two main parts: the self-attention   
  enhancement  module, and the  
detection module.
An initial illumination map $\tilde{t}$  is  estimated
by using the BCP \cite{wang2013automatic}. Then  the 
 encoder-decoder network      is used to obtain a final illumination map $t$ an with the unsupervised loss function $L_{BCP}$.  Moreover,  the  self-attention map  $T_{att}$  sourced from the infrared modal   is applied in the enhancement network to obtain more informative representation focused  on pedestrian.

\subsection{Motivation}

Existing  Multispectral Pedestrian
Detection  methods aim at   effectively extracting information from thermal images and visible light images to improve detection quality  ,but still meet some challenges:

\par (1)
From existing Multispectral Pedestrian
Detection  methods, we can
 observe that many of them  attempt to fuse visible light images and thermal images \cite{liu2016multispectral,zhou2020improving}. However, such fusion might reduce the quality of the results  for the  imbalanced information between thermal images and visible light images.
\par(2)
There are many methods  \cite{he2010single,guo2016lime} of nighttime enhancement that can be useful for  multispectral pedestrian detection. However, better obtaining task-oriented  enhancement of RGB images,  remains an interesting challenge.

To address the issues mentioned above, we propose 
  framework as shown in Fig. 
 .  Firstly, as  pedestrians have  temperature, their silhouettes are  visible in thermal images. Using the V-channel (brightness) of the HSV image of the thermal image as the attention map, the unsupervised auto-encoder for visible light images can be activated layer by layer to focus more on pedestrian features. Secondly,  we use enhanced networks and unsupervised bright channel prior algorithms for light compensation in low light images. In addition,  we combine image enhancement and object detection through a unified framework to obtain enhancement that benefits detection tasks. 
 
\subsection{Bright  Channel Prior Algorithm  }

 \begin{figure*}[!t] 
    \centering %
    \includegraphics[width=17cm]{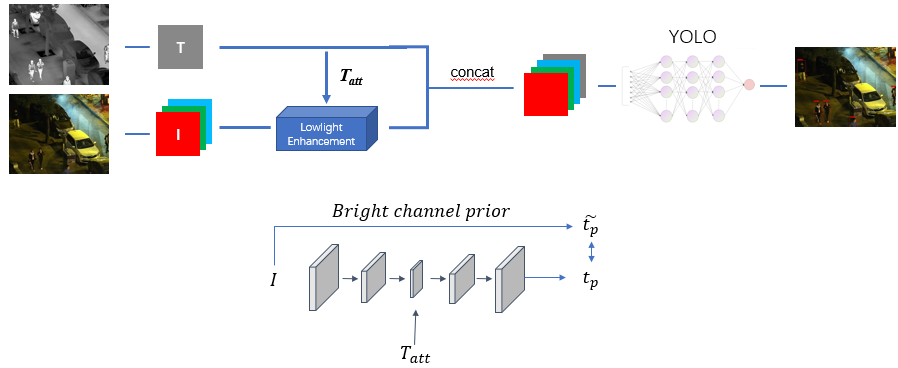}
    \caption{The framework of the proposed method.}
    \label{framework1}
\end{figure*}

The Brightness Correction Processing (BCP) algorithm \cite{wang2013automatic} was  a means of adjusting local exposure within an image. This approach is derived from the Dark Channel Prior (DCP) [33], which was initially created to forecast the transmission map of a given hazy input image. We employ the BCP algorithm to predict an initial illumination map by  introducing the unsupervised loss in our proposed enhancement network. For the BCP low-light enhancement module, we define it as follows:
\begin{equation}
\label{bcp}
I_p = t_p J_p + (1 - t_p)A,
\end{equation}
where  $I_p$  represents an observed low-light RGB image ,  $J_p \in \mathrm{R^3}$ is an enhanced  image, $t_p \in \mathrm{R}$   is an illumination map and  $A\in \mathrm{R^3}$ indicates an environment light.   Assuming that  $A$ is 

 is known and the illumination map $\tilde{t}$, 
, remains constant within $\Omega(p)$ which is the patch centered at the pixel  $p$ 
,we can derive the following equation by applying the max operator to both sides of Equation (1) :
\begin{equation}
\tilde{t_p}= 1-\mathop{max}\limits_{c,q} (\frac{1-I_q^c}{1-A^c}) 
\end{equation}

The darkest pixel in an image can be used to represent the environment light. Nonetheless, in a real-world setting, the darkest pixel might correspond to a shadow or a dark object. Therefore, following the approach in  \cite{lee2020unsupervised}, a set of the darkest 0.1\% pixels (referred to as K) is used to compute A in the brightest channel of I (represented as $\max_{c\in r,g,b} I^c$):
\begin{equation}
A = \frac{1}{|K|}\sum_{p\in K}I_p 
\end{equation}

\subsection{Unsupervised  Cross-Modal Illumination Map Estimation}

\paragraph{Unsupervised BCP Loss} Given the   $t_p$  obtained through the   
 encoder-decoder network under the   supervision  of $\tilde{t_p}$  ,  we have the enhanced image $J_p$ according to the Eq. \ref{bcp}:
 \begin{equation}
J_p= \frac{I_p-A}{t_p} +A    
 \end{equation}

Inspired by \cite{wang2013automatic},   considering 
 As Eq. \ref{bcp}  is conceptually similar to image matting, where an output image is expressed as a linear combination of foreground and background, we  decide to adopt the regularizer used in soft matting . 
Overall, the  unsupervised BCP loss is formulated as:
\begin{equation}
    L_{BCP} = \frac{1}{N} \sum_P \{ (t_p-\tilde{t_p})^2 + \lambda \sum_{i,j \in \Phi(P)} w_{ij}(t_i-t_j)^2   \},
\end{equation}
where  N is the total number of pixels, $\lambda$ is a weight parameter,  $w_{ij}$   is value of  the  Laplacian matrix that calculates the affinity between $I_i$ and $I_j$, $\Phi(P)$ represents 3 × 3 patch around center pixel $p$. 

\paragraph{Cross-Modal Spatial 
 Attention Module}
The single  unsupervised light enhancement could not  
guarantee the  output image    beneficial to pedestrian detection. From Fig. \ref{framework1}  we can observe that 
pedestrians have  temperature,  contour  of target to be detected is  visible in thermal images. 
Therefore,   we use the brightness channel (V-channel) from the HSV image of the thermal image as a cross-modal  attention map: 
\begin{equation}
T_{att} = (T^V)^\gamma,
\end{equation}
where $\gamma$ is a parameter that
controls the curvature of the attention map.

  $T_{att}$  is multiplied
to convolutional activations of all layers in the enhancement
network E, as shown in Fig. 2. This enables  for adapting the  proposed network accordingly depending on the  presence of  pedestrians  to  capture more  needed 
 details and properly   ignore the background  of   RGB images.


\begin{figure*}[!t]
	\centering
	\begin{subfigure}{ 0.245\linewidth}
		\centering
		\includegraphics[width=1\linewidth]{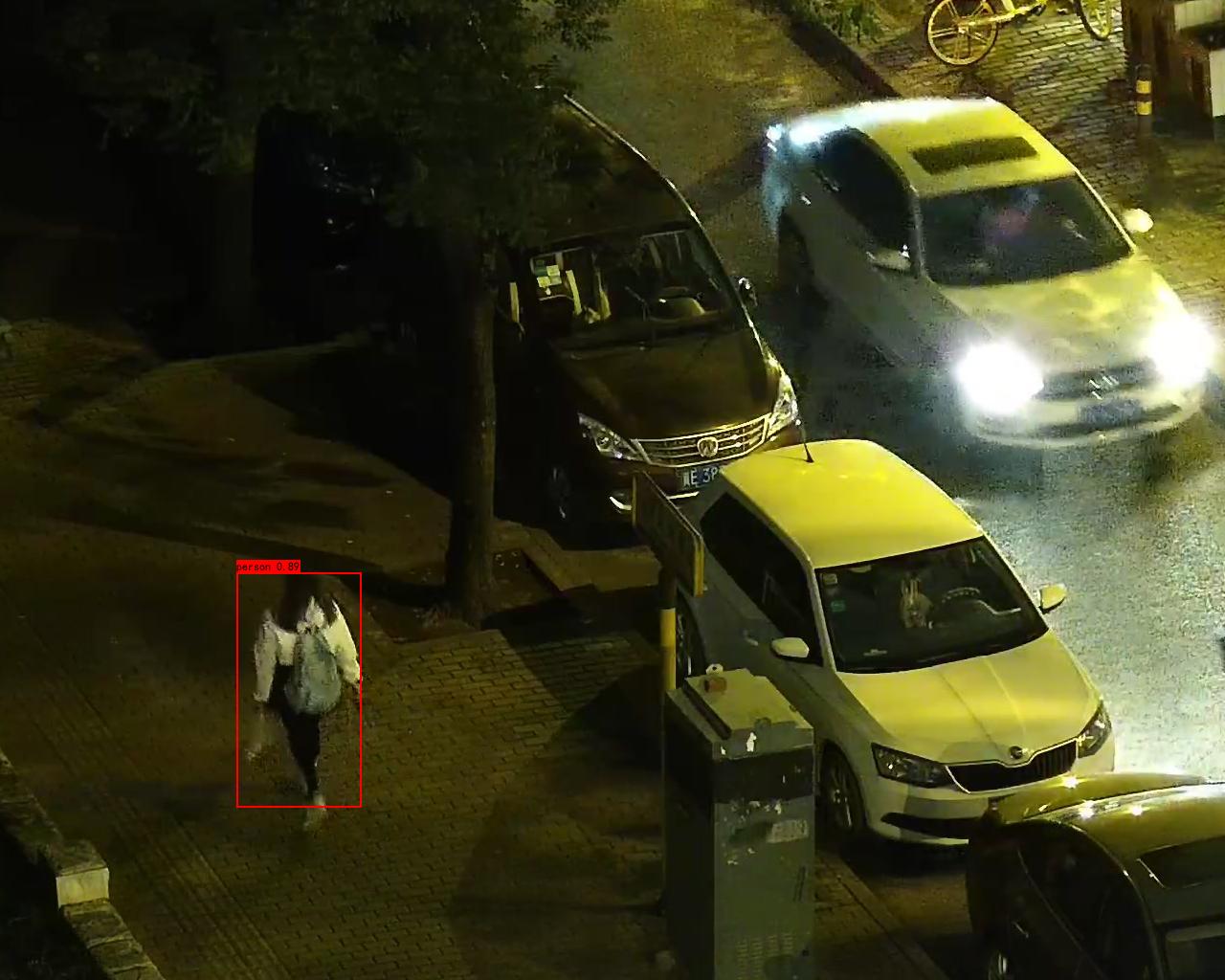}

	\end{subfigure}
 	\centering
	\begin{subfigure}{ 0.245\linewidth}
		\centering
		\includegraphics[width=1\linewidth]{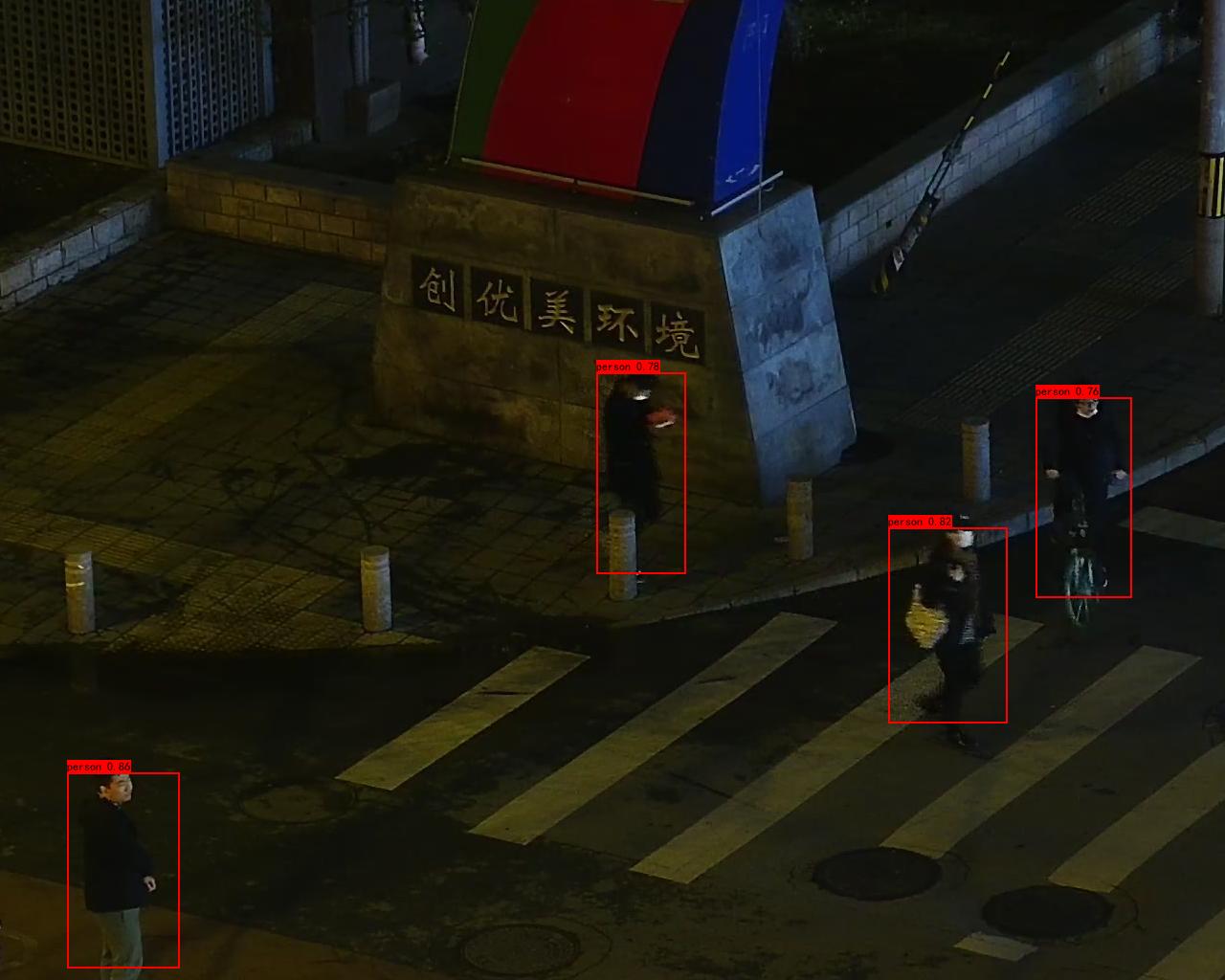}

	\end{subfigure}
 	\centering
	\begin{subfigure}{ 0.245\linewidth}
		\centering
		\includegraphics[width=1\linewidth]{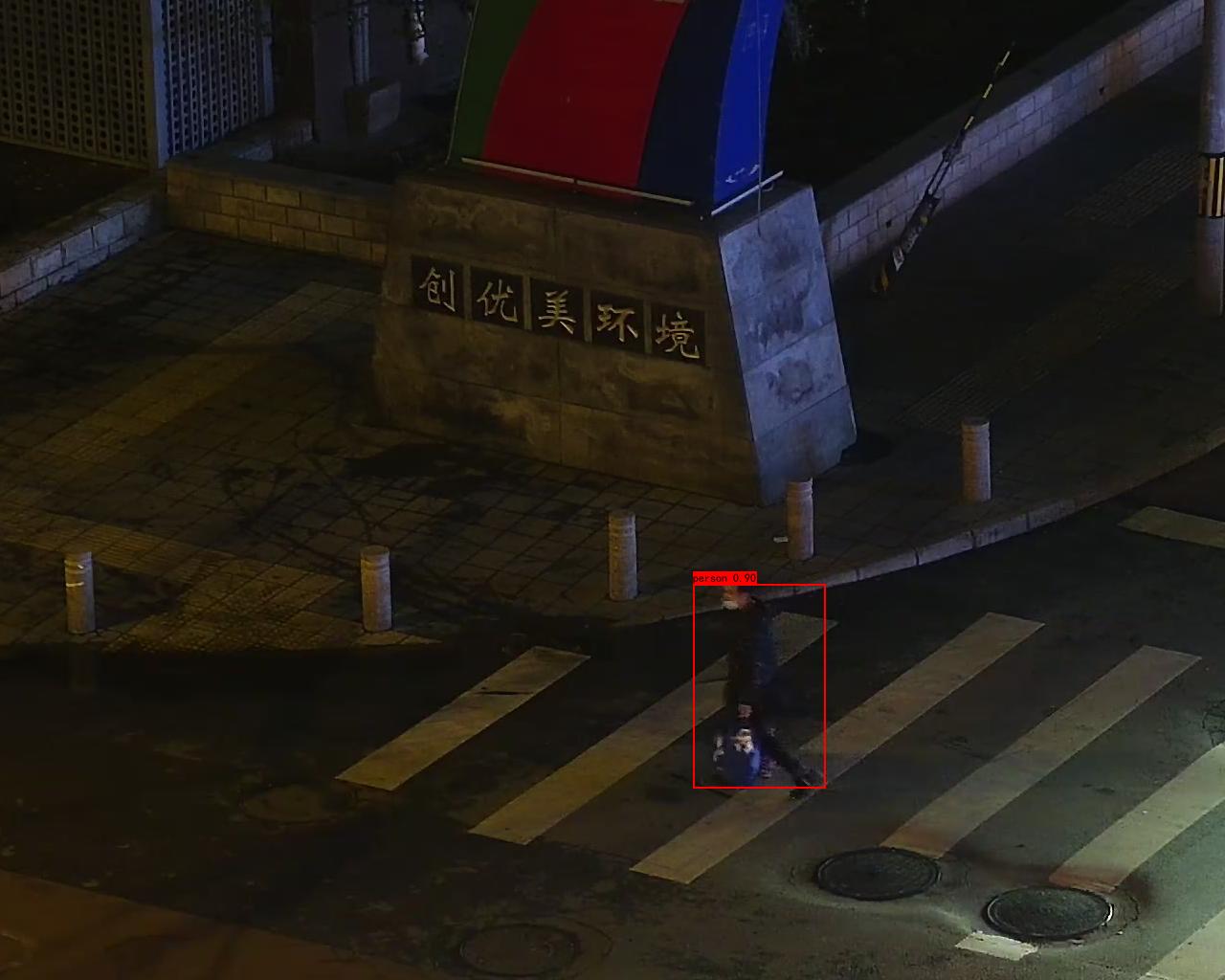}
 
	\end{subfigure}
	\centering
	\begin{subfigure}{ 0.245\linewidth}
		\centering
		\includegraphics[width=1\linewidth]{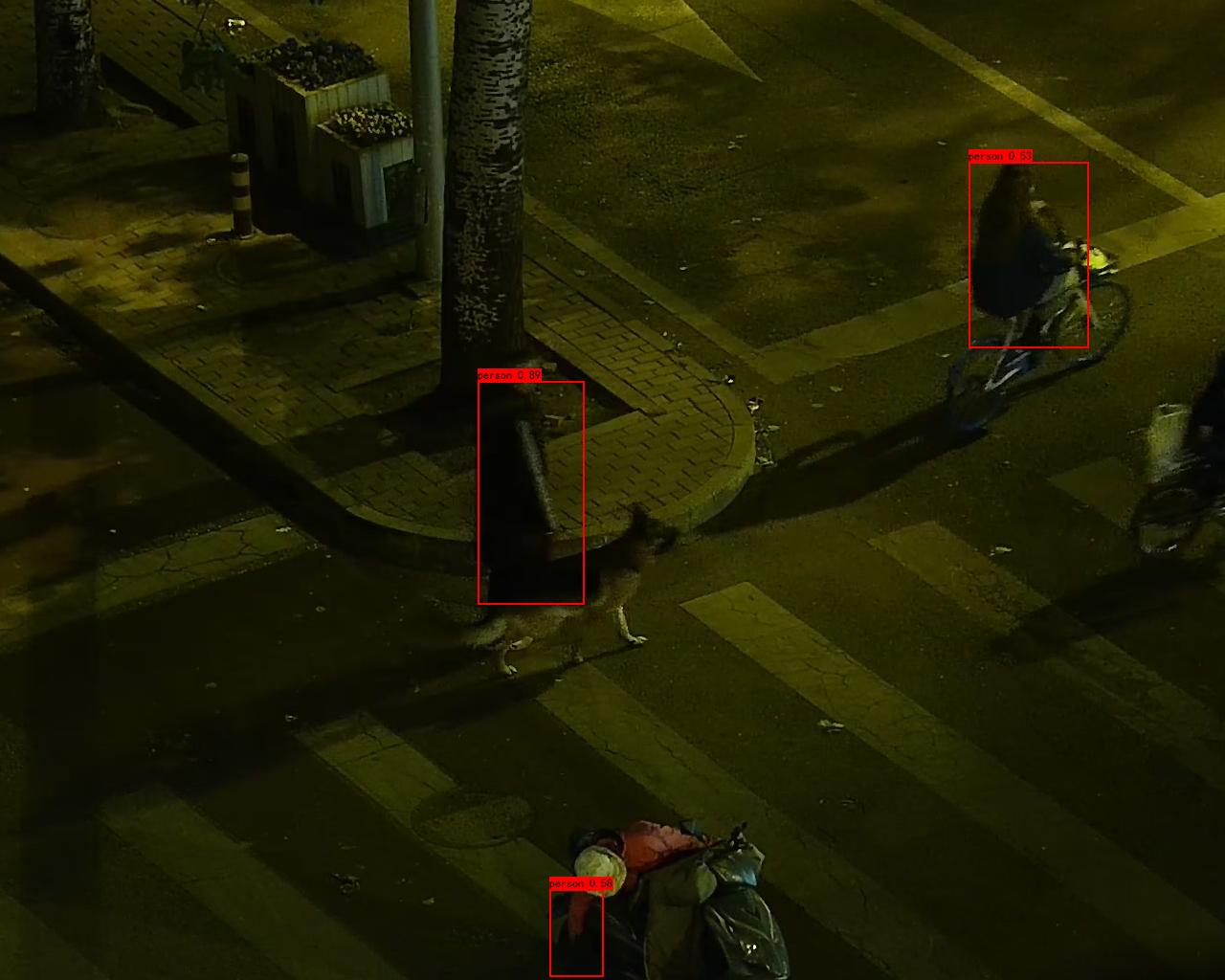}
 
	\end{subfigure}

	\centering
	\begin{subfigure}{ 0.245\linewidth}
		\centering
		\includegraphics[width=1\linewidth]{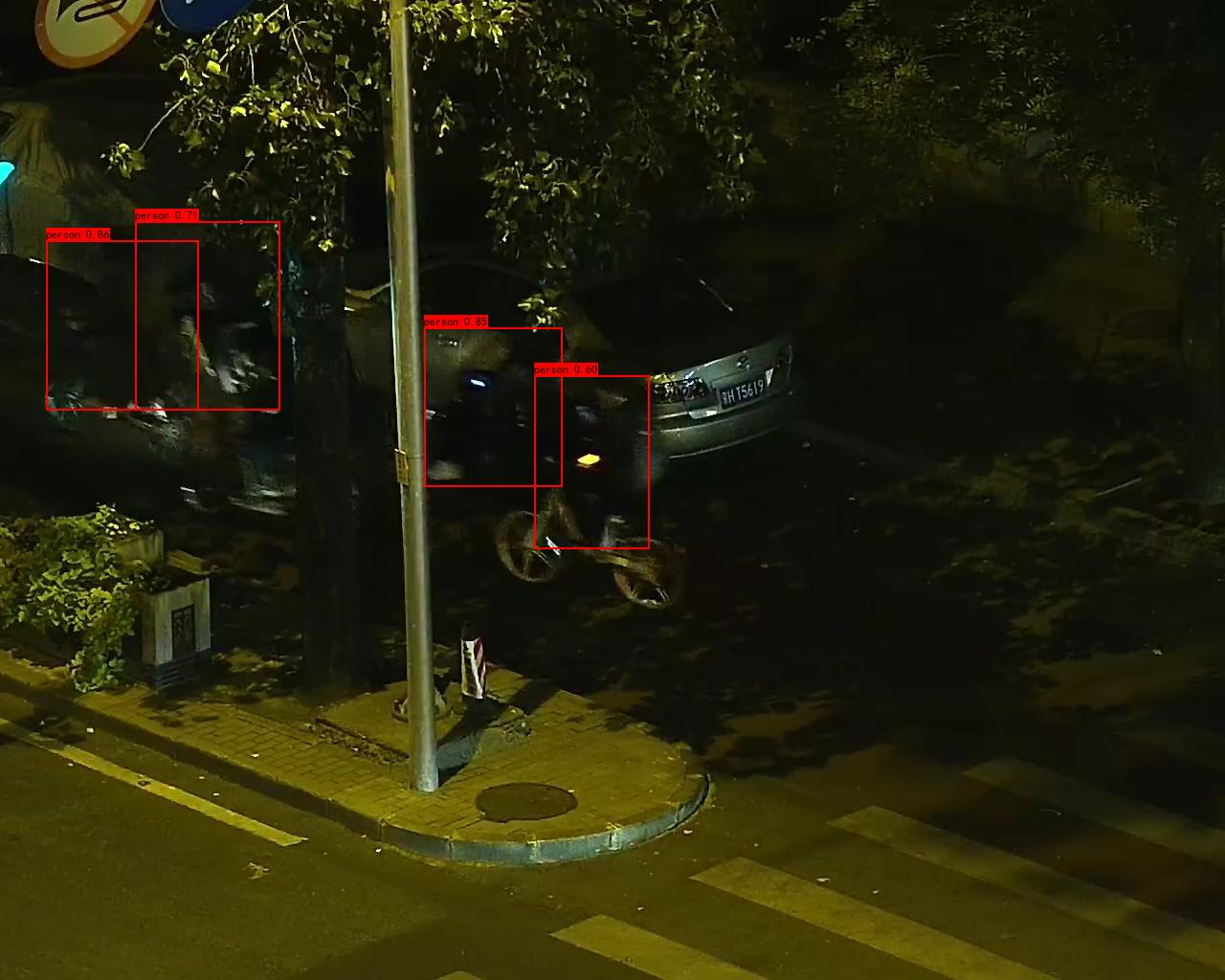}
 
	\end{subfigure}
 	\centering
	\begin{subfigure}{ 0.245\linewidth}
		\centering
		\includegraphics[width=1\linewidth]{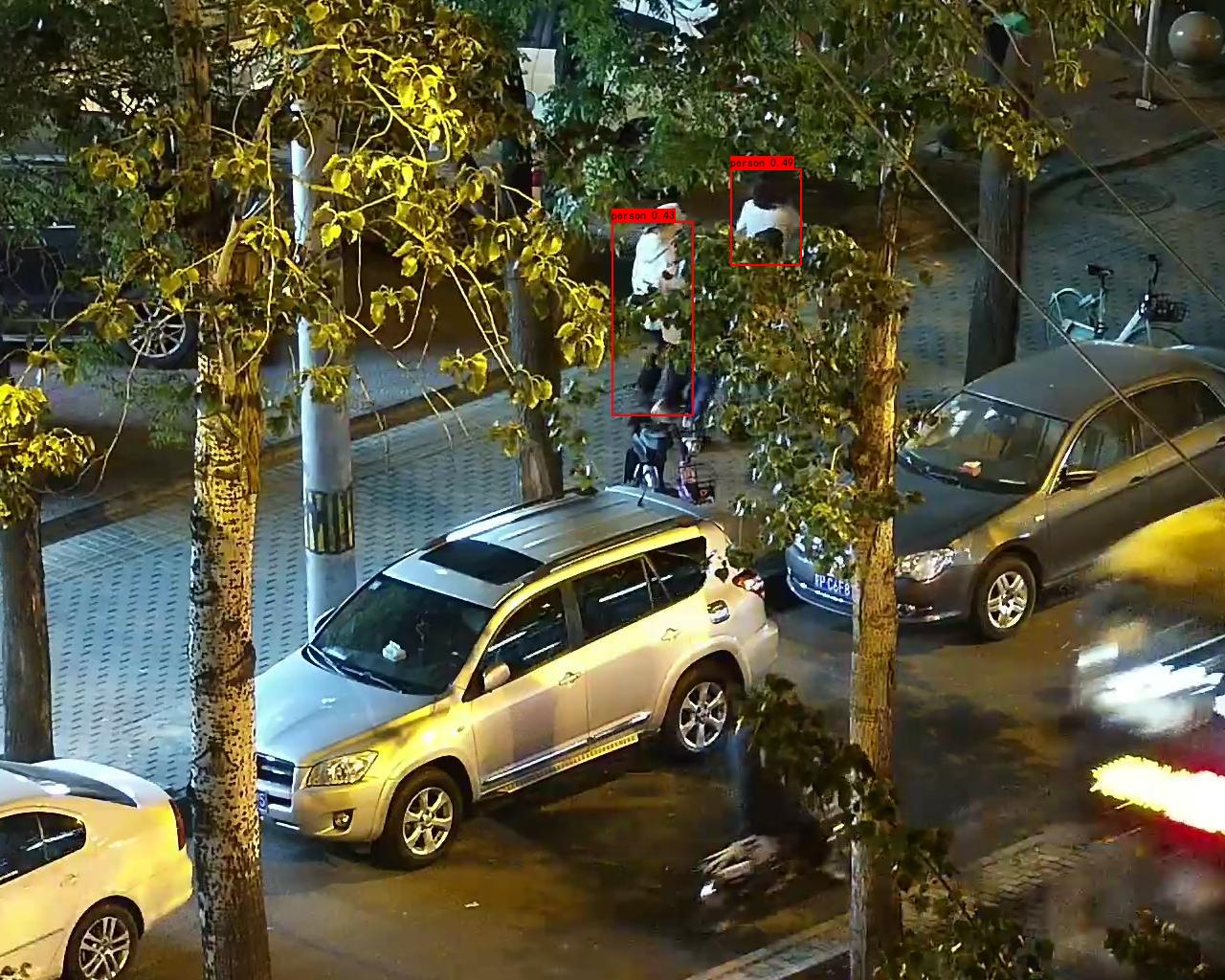}

	\end{subfigure}
 	\centering
	\begin{subfigure}{ 0.245\linewidth}
		\centering
		\includegraphics[width=1\linewidth]{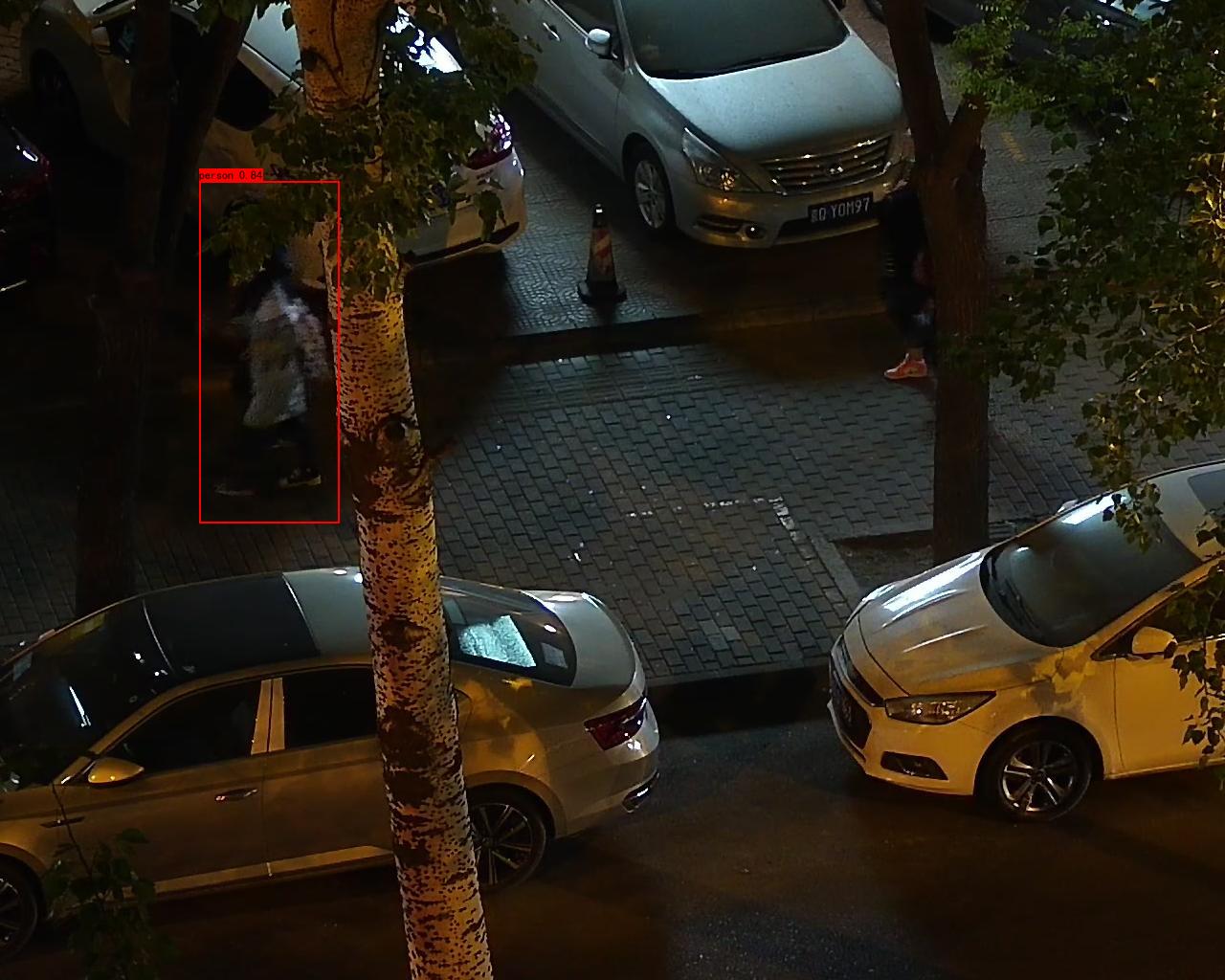}
 
	\end{subfigure}
	\centering
	\begin{subfigure}{ 0.245\linewidth}
		\centering
		\includegraphics[width=1\linewidth]{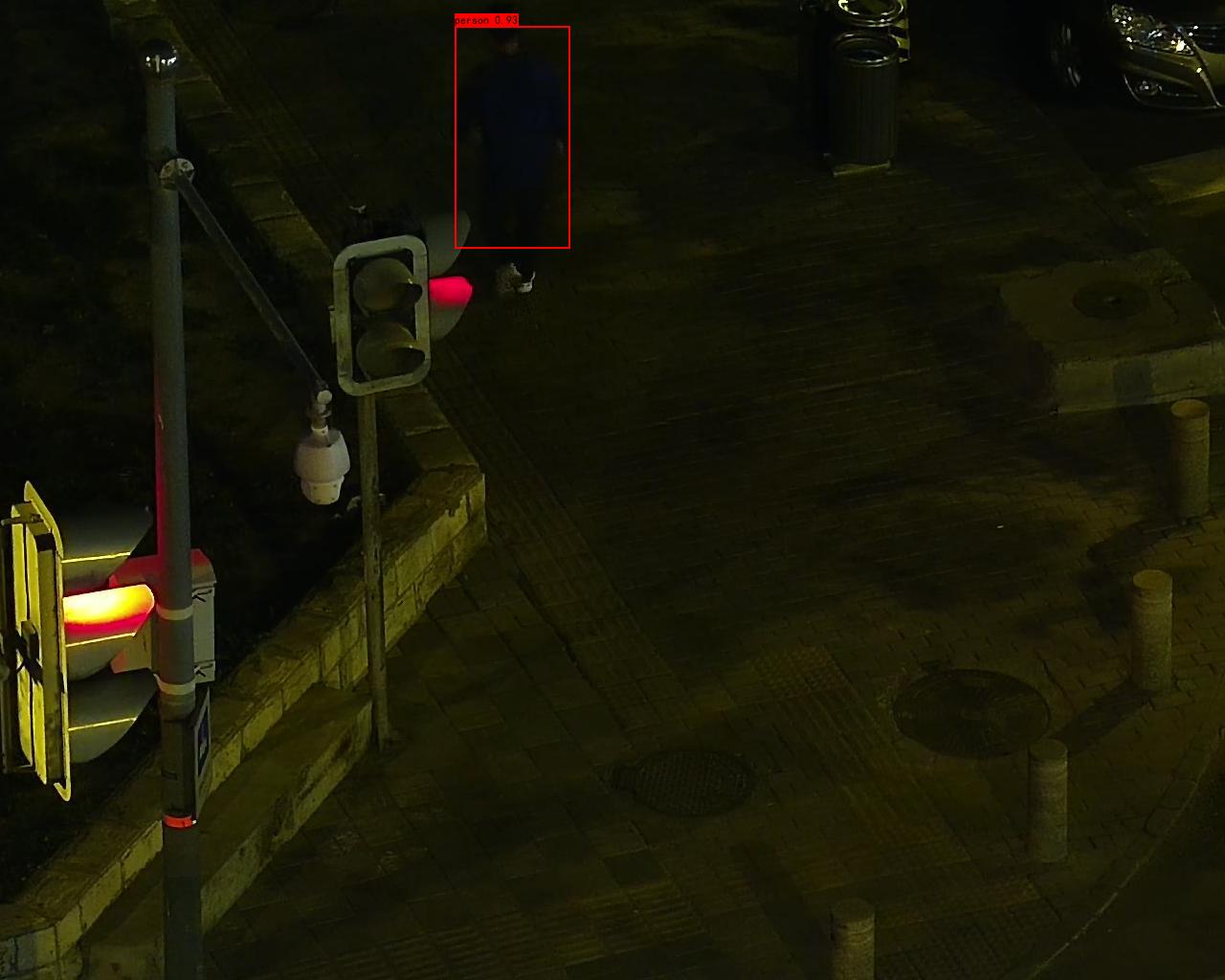}
 
	\end{subfigure}
 
 	\caption{Display of detection results on validation set.}
	\label{dis}
\end{figure*}

\subsection{YOLO-based Detection Module}

YOLO-v4, also known as You Only Look Once version 4, is a state-of-the-art real-time object detection system  widely used in computer vision and machine learning research. The YOLO-v4 object detection system builds on the success of its predecessors YOLO and YOLOv3, and is known for its improved detection accuracy and speed.
We  combine our enhancement network with YOLO-v4 and use  the concat of enchanced image and  infrared image as the input of  YOLO detection module. The total loss function $L$ is as follows :

\begin{equation}
     L =  L_{BCP} + \beta L_{YOLO}
\end{equation}
Combining detection and augmentation loss for training is beneficial for obtaining images for the detection task. 
  Moreover, we establish a unified framework consisting of   enhancement network and detection   network  for multispectral pedestrian detection



\section{Experiments}

\noindent{\textbf{Compared Methods.}}
We compare our Method with  Yolo-v4. Moreover,  we explored the four-channel version of Yolo-v4, where the thermal image and visible light image are concatenated as the input of YOLO. We report the accuracy, recall and mAP of the detection results. The results are shown in Table \ref{datasets}.  

\begin{table}[!h]
\caption{The results of all methods on LLVIP.}
\centering 
{

\begin{tabular}{c|ccc}

\hline
Method  & Yolo-v4-3c&  Yolo-v4-4c  & ours  \\ \hline
accuracy &0.924   & 0.598 &  0.947\\
recall & 0.601 & 0.571   &   0.647\\
 mAP & 0.671  & 0.509&0.740  \\
 \hline
\end{tabular}
}

 \label{datasets}
\centering 
\end{table}
\noindent\textbf{Dataset.}
We include use RGB and Thermal  
dataset LLVIP \cite{llvip} to test the performance of our methods. LLVIP (low light visible image person) dataset is a benchmark dataset designed for low-light visible image person detection tasks. The dataset consists of over 10,000 person instances captured in low-light scenarios, with large variations in poses, scales, occlusions, and illuminations. Moreover, the dataset includes annotations of bounding boxes and class labels for each person instance. LLVIP dataset covers a range of applications such as video surveillance, nighttime photography, and autonomous driving. The dataset is divided into two subsets, with 60\% of the data in the training set and the remaining 40\% in the test set. The LLVIP dataset contributes to the computer vision community by presenting a new and challenging dataset for low-light visible image person detection.

\noindent{\textbf{ Display of dection results.}}
To further show the effectiveness of the proposed method,  we display the detection results  on validation set of LLVIP as shown in Fig. \ref{dis}. 

\noindent{\textbf{Implementation.}} The experiments were conducted on a Windows PC equipped with an Intel (R) Core (TM) i5-12600K CPU @ 3.69 GHz, 32 GB RAM, and a GeForce RTX 3070ti GPU with 8 GB of cache memory.
\section{Conclusion}
In this paper, we proposed an attention-based feature fusion algorithm for multimodal pedestrian detection in low light conditions. The proposed algorithm utilizes the V-channel of the HSV image and unsupervised auto-encoder for attention and feature extraction, respectively. Our approach addresses the challenge of light compensation and improves the object detection accuracy in low light conditions. Additionally, integrating image enhancement and detection in a unified framework provides enhancement benefits to detection tasks. Our contributions offer promising results for autonomous driving and different surveillance scenarios. The proposed approach presents a novel and robust solution for pedestrian detection in low light conditions, which can be extensible to other multimodal pedestrian detection applications. We will further supplement the experiments in future work.


\bibliographystyle{plain}
\bibliography{egbib}
 
\end{document}